ARTICLE

# DT-NeRF: A Diffusion and Transformer-Based Optimization Approach for Neural Radiance Fields in 3D Reconstruction

**Bo Liu[1,+], Runlong Li[2,+], Li Zhou[3,*] and Yan Zhou[4]**

[1] College of Computer Sciences, Northeastern University, Cupertino, CA, 95014, USA
[2] Department of Electrical Engineering and Computer Science, University of California, Irvine, Moreno Valley, CA, 92555, USA
[3] Desautels Faculty of Management, McGill University, Montréal, 27708, Canada.
[4] Department of Mathematics, Northeastern University, San Jose, CA, 95131, USA
[+] Co-First Author

**Abstract**

This paper proposes a Diffusion Model-Optimized Neural Radiance Field (DT-NeRF) method, aimed at enhancing detail recovery and multi-view consistency in 3D scene reconstruction. By combining diffusion models with Transformers, DT-NeRF effectively restores details under sparse viewpoints and maintains high accuracy in complex geometric scenes. Experimental results demonstrate that DT-NeRF significantly outperforms traditional NeRF and other state-of-the-art methods on the Matterport3D and ShapeNet datasets, particularly in metrics such as PSNR, SSIM, Chamfer Distance, and Fidelity. Ablation experiments further confirm the critical role of the diffusion and Transformer modules in the model's performance, with the removal of either module leading to a decline in performance. The design of DT-NeRF showcases the synergistic effect between modules, providing an efficient and accurate solution for 3D scene reconstruction. Future research may focus on further optimizing the model, exploring more advanced generative models and network architectures to enhance its performance in large-scale dynamic scenes.

## 1 Introduction

Neural Radiance Fields (NeRF) have emerged as a novel 3D scene reconstruction technology, attracting widespread attention in the fields of computer vision and graphics in recent years. The core concept of NeRF is to utilize a multi-layer perceptron (MLP) to learn the propagation of light in a 3D scene from different viewpoints, enabling high-quality 3D reconstruction. NeRF is capable of generating intricate lighting and texture details, particularly making significant advancements in modeling complex scenes and achieving photorealistic rendering[1]. Despite its strong performance in 3D reconstruction, NeRF still faces challenges when dealing with sparse viewpoints and complex geometries[2]. Due to its strong reliance on viewpoint information, the reconstruction quality often suffers when the viewpoints are insufficient, leading to loss of details or poor consistency[3][4].

As a result, traditional NeRF has some inherent limitations, especially in scenarios with sparse viewpoints and complex geometries, where reconstruction quality and efficiency are often constrained[5]. Specifically, NeRF struggles to effectively recover details when there are fewer viewpoints, and maintaining scene consistency becomes challenging. Moreover, NeRF's computational cost is high, particularly when handling complex geometries and large-scale scenes, where training time and inference speed become bottlenecks. While NeRF can generate high-quality images via volumetric rendering, it still exhibits certain limitations in modeling long-range dependencies and reconstructing complex structures, especially in scenes with intricate geometric details, where the traditional MLP architecture may not fully capture all spatial relationships[6].

To address these issues, recent studies have increasingly integrated other advanced deep learning techniques with NeRF to enhance its performance in complex scenes[7]. Some methods introduce generative models to gradually optimize the details and quality of images. These generative models have shown significant success in image restoration and enhancement, particularly in recovering details and improving generated image quality. For example, some denoising-based generative models can iteratively remove noise at each step, optimizing the final image's quality and consistency. Additionally, another class of methods introduces global feature aggregation strategies to effectively handle long-range dependencies in 3D point cloud data, enabling the model to better capture spatial relationships in complex geometric scenes, thereby improving the ability to reconstruct geometric details[8]. By combining these methods, it is expected that NeRF's detail recovery under sparse viewpoints, multi-view consistency, and the optimization of 3D scene geometries will be significantly enhanced[9]. The main contributions of this paper are as follows:

- We introduce a diffusion model into the training process of NeRF, which generates latent features that effectively compensate for the lack of viewpoints in sparse-view scenarios. This aids in detail recovery and improves image quality, addressing the limitations of traditional NeRF in these conditions.

- We embed a Transformer into the rendering process of NeRF, replacing the traditional MLP structure with self-attention mechanisms. This improvement enhances the model's ability to capture long-range dependencies in 3D scenes, which significantly boosts the accuracy of geometric modeling and detail reconstruction.

- By combining these two advanced techniques—diffusion models and Transformers—we propose an efficient joint optimization framework that results in significant improvements in image quality, geometric accuracy, and multi-view consistency, making the method particularly effective for complex 3D scene reconstruction tasks.

The structure of this paper is as follows: Section 2 reviews related works, including the basic principles and limitations of NeRF, applications of diffusion models in image generation, and relevant uses of Transformers in computer vision. Section 3 provides a detailed description of the proposed DT-NeRF model, including the overall architecture, specific designs of the diffusion model module and the Transformer module. Section 4 presents the experimental setup, datasets, evaluation metrics, and validation of our model's effectiveness through comparative and ablation experiments. Finally, Section 5 summarizes the contributions of this paper and outlines future research directions.

## 2 Related Work

### 2.1 Application of Traditional Methods in Image Generation and 3D Scene Reconstruction

In recent years, with the rapid development of computer vision technologies, numerous methods have been proposed to address the challenges in image generation and 3D scene reconstruction[10]. Traditional image generation methods, such as multi-view stereo (MVS)-based techniques, restore the 3D geometric information of a scene by matching and fusing images from multiple viewpoints[11]. However, these methods rely on strong dependencies between viewpoints and often face performance bottlenecks in scenarios with sparse viewpoints or complex reconstruction details. Another class of methods is based on volumetric rendering, using ray tracing for scene reconstruction, such as employing the Marching Cubes algorithm and voxel grids for spatial partitioning and modeling[12]. While this approach has certain advantages in efficiently reconstructing scene structures, it incurs high computational overhead[13]. Additionally, some methods use image-based reconstruction techniques, such as structured light scanning and stereo vision, which utilize image segmentation and depth estimation technologies for rapid 3D scene reconstruction. Although this method is relatively simple, it still has limitations in terms of detail representation and geometric complexity[14]. Furthermore, SLAM (Simultaneous Localization and Mapping) technology, as a real-time 3D mapping method, enables dynamic scene reconstruction using information obtained from cameras or other sensors, but its performance in large-scale environments remains constrained by hardware limitations[15]. Recently, the deep learning-based NeRF method has achieved significant results in high-quality 3D reconstruction by modeling light propagation[16]. However, its dependence on numerous viewpoints and computational resources limits its application in scenarios with sparse viewpoints and complex scenes[17].Recent advancements in diffusion models, such as Stable Diffusion and DDPM, have shown promise in overcoming these challenges, particularly for sparse-view reconstruction. Additionally, Vision Transformers (ViTs) have been successfully applied to 3D tasks, improving the capture of long-range dependencies in complex scenes. These advancements support the motivation for combining diffusion models and Transformers in DT-NeRF[18][19].

Compared to these traditional methods, this paper introduces diffusion models and Transformers to optimize NeRF, aiming to address issues related to detail loss under sparse viewpoints and deficiencies in modeling complex geometries. Unlike traditional methods that rely on explicit geometric modeling or image matching, our approach enhances detail restoration through generative models and improves the capture of geometric and spatial information using self-attention mechanisms in deep learning, thereby improving both the effectiveness and efficiency of 3D reconstruction.

## 2.2 Innovative Application of Deep Learning in 3D Scene Reconstruction

In recent years, with the continuous development of deep learning technologies, many studies have gradually integrated deep neural networks with 3D scene reconstruction[20][21], especially with significant advancements in the application of Neural Radiance Fields (NeRF) in this field. Many NeRF-based optimization methods have attempted to improve the model's performance in complex scenes. For example, NeRF-W (NeRF with Weakly Supervised Learning) handles data from unlabelled viewpoints through weakly supervised learning, significantly improving NeRF's reconstruction capability in sparse viewpoint scenarios[22]. Additionally, FastNeRF optimizes the computational process of NeRF by utilizing hierarchical networks and acceleration techniques, improving training speed and inference efficiency, thus making real-time applications feasible[23]. However, while these methods enhance the model's efficiency and detail recovery capabilities, they still face certain limitations in handling complex lighting, long-range dependencies, and geometric structure modeling, particularly in large-scale and dynamic scenes, where detail recovery and consistency maintenance are not ideal. Meanwhile, another line of research has attempted to combine Generative Adversarial Networks (GANs) with NeRF to enhance the details and realism of reconstructed images. For instance, NeRF-GAN combines the generative capabilities of GANs with NeRF's volumetric rendering, improving scene detail and texture representation[24]. However, it still faces challenges in maintaining multi-view consistency, particularly when processing dynamic scenes, where the generated images may exhibit inconsistencies. Other studies, such as PointNeRF and NeRF-T, have optimized geometric modeling and global context capturing by incorporating point clouds and Transformer architectures, thereby improving the accuracy of 3D reconstruction[25][26]. However, these methods still rely on a large amount of viewpoint information and involve high computational complexity. Thus, despite progress in some areas, existing methods continue to face challenges in detail recovery, geometric modeling, and computational efficiency[27]. The computational cost of NeRF-W and the instability of GAN-based NeRF methods further highlight the need for a more efficient and stable solution, which is addressed by the DT-NeRF model.

In contrast to these methods, the DT-NeRF model proposed in this paper combines diffusion models with Transformers to further enhance detail recovery and geometric modeling capabilities while retaining the advantages of traditional NeRF. We optimize the training process through latent features generated by the diffusion model and enhance global context modeling using Transformers. This approach addresses the shortcomings in sparse viewpoints and complex geometric modeling while also optimizing computational efficiency.

## 3 Methodology

### 3.1 Overall Model Architecture

The DT-NeRF model combines diffusion models and Transformers to optimize the performance of NeRF in complex 3D scene reconstruction. Figure 1 illustrates the architecture of DT-NeRF, where the diffusion model, Transformer, and NeRF decoder are closely integrated to form an innovative end-to-end 3D reconstruction framework. This framework first generates latent features through the diffusion model, which are then used as conditional inputs for the NeRF decoder. The Transformer module optimizes the input data through a self-attention mechanism, ultimately enhancing the accuracy of scene geometric modeling and detail recovery. The design of the overall architecture enables DT-NeRF to effectively address the challenges of sparse viewpoints and complex geometric reconstruction.

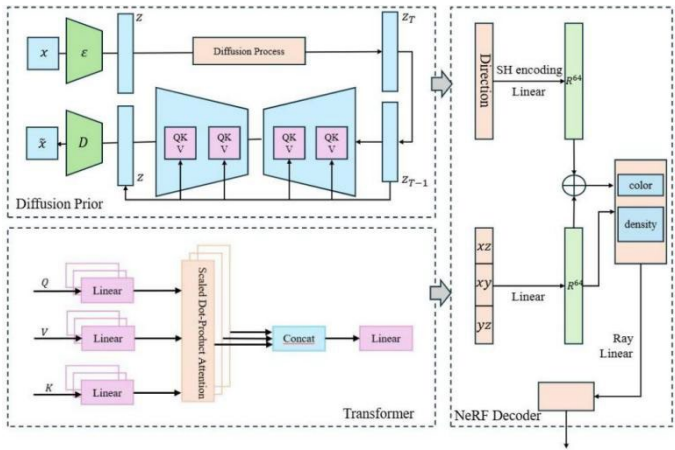

**Figure 1.** Architecture of the DT-NeRF Model.

In DT-NeRF, the role of the diffusion model is critical. The diffusion model progressively denoises the initial noisy image to transform it into latent features. During this process, the diffusion model not only effectively enhances image details but also generates rich latent features during training, which are passed as conditional inputs to the NeRF decoder. In this way, the diffusion model provides more refined input data for NeRF while addressing the issue of information scarcity caused by sparse viewpoints. Particularly when the viewpoints are limited, the diffusion model compensates for the shortcomings of traditional NeRF methods in such scenarios by enhancing the latent features of the input, thus improving the detail and consistency of the reconstruction.

Complementing this is the introduction of the Transformer module. The primary role of the Transformer in DT-NeRF is to optimize NeRF's geometric modeling capabilities. By introducing the self-attention mechanism, the Transformer module enables global context modeling of 3D point cloud data, capturing long-range dependencies and details in complex geometric structures[28]. This process significantly enhances the model's geometric modeling capabilities in complex scenes, particularly when dealing with scenes with highly intricate structures. The Transformer ensures that geometric details in the reconstruction process are more accurate by focusing on the relationships between different 3D points. Additionally, the Transformer-optimized feature inputs effectively improve the performance of the NeRF decoder, resulting in finer and more consistent reconstructed images.

The design of the DT-NeRF model innovatively combines the advantages of both the diffusion model and the Transformer. The diffusion model generates high-quality latent features, enhancing NeRF's input and compensating for the lack of viewpoints in sparse-view scenarios. The Transformer improves geometric reconstruction and global context modeling by capturing long-range dependencies, boosting accuracy in complex scenes. Both the Diffusion and Transformer models are trained jointly within a unified optimization framework, sharing the same loss function. These modules are not pretrained; instead, they are optimized simultaneously during training through backpropagation, allowing them to learn collaboratively and complement each other. Gradients are propagated through both models during backpropagation, ensuring that both components are updated together. The computational complexity of training DT-NeRF increases with the addition of the diffusion model and Transformer components, which require more parameters and more computational resources compared to traditional NeRF. The training time is also affected by the need to jointly optimize both modules. Despite the increased complexity, the enhanced accuracy and detail recovery capabilities justify the trade-off in computational cost. Through this joint optimization, DT-NeRF not only improves the quality of the reconstruction results but also enhances its performance across various scenes, particularly in handling sparse viewpoints and complex geometries, delivering higher-quality outputs in 3D scene reconstruction[29].

## 3.2 Detail restoration and feature generation

The diffusion model module in the DT-NeRF model plays a crucial role in enhancing the detail recovery and enhancement capabilities of 3D scene reconstruction[30]. Figure 2 illustrates the overall architecture of this module, which includes the entire process of noise addition, reverse denoising, and latent feature generation. The goal of the diffusion model is to progressively transform the image into a noisy image and then, through the reverse process, recover high-quality latent features. These latent features are ultimately used as conditional inputs for the NeRF decoder, thereby improving the detail representation in 3D scene reconstruction. In this process, the diffusion model learns, through a deep neural network, how to recover details from noise that are as close as possible to the original image, providing effective support for NeRF's training.

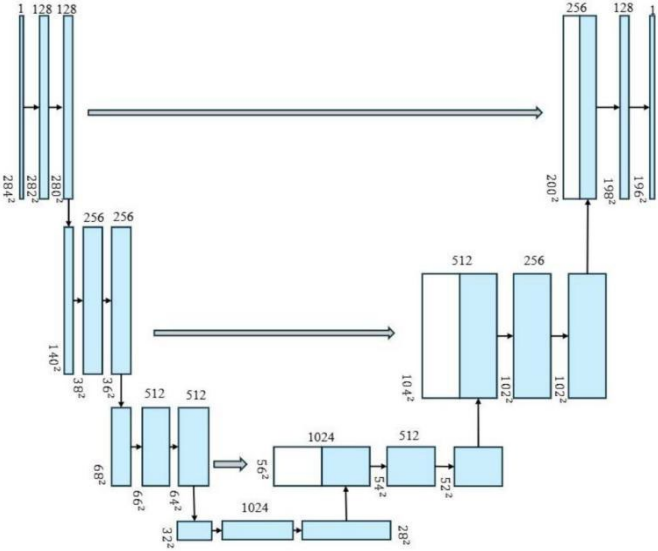

**Figure 2.** Architecture of the Diffusion Model for Detail Recovery and Latent Feature Generation in DT-NeRF, which generates high-quality latent features to restore details in sparse-view scenarios. The components show key stages of the diffusion process, including input processing, feature generation, and integration with NeRF input, enhancing multi-view consistency and geometric accuracy in complex scenes.

The first step in the diffusion process is to add noise to the input image. Starting with the original image $x_0$, we generate a noisy image $x_t$ through a series of noise steps. $\alpha_t$ is a coefficient that adjusts the noise intensity, and $\epsilon_t$ is the noise sampled from a standard normal distribution. As the time step t increases, the image $x_t$ gradually becomes blurred, eventually approaching pure noise:

$$x_t = \sqrt{\alpha_t} x_0 + \sqrt{1 - \alpha_t} \epsilon_t \qquad (1)$$

In the denoising stage, the diffusion model uses a deep neural network $f_\theta(x_t, t)$ to progressively restore the details of the image. The input to the network is the noisy image $x_t$ and the time step t, and the output is the denoised latent image $\hat{x}_0$, which is the recovered clear image:

$$\hat{x}_0 = f_\theta(x_t, t) \qquad (2)$$

During the training process, the neural network optimizes the network parameters θ to minimize the difference between the recovered image and the real image. In each iteration, the diffusion model optimizes the denoising process through contrastive loss, allowing the model to learn how to recover high-quality latent features from the noisy image. The recovered latent feature $\hat{x}_0$ is then fed into the NeRF decoder module, providing the conditional input for 3D scene reconstruction:

$$L = E_{x_0, t} \left[ \| x_0 - f_\theta(x_t, t) \|_2 \right] \qquad (3)$$

The diffusion model plays an important role not only in image recovery but also in providing reliable features for multi-view scene reconstruction. During the training process, the diffusion model generates high-quality latent features, addressing the issue of detail loss in sparse viewpoint scenarios and effectively improving multi-view consistency. Particularly in scenes with insufficient viewpoints, the diffusion model provides enough information to NeRF, enhancing the model's ability to recover details, thereby ensuring high-quality output for scene reconstruction. In this way, the diffusion model module plays a key role in DT-NeRF, effectively improving the performance and accuracy of scene reconstruction.

### 3.3 Geometric modeling and long-range dependencies

The Transformer module in DT-NeRF plays a crucial role in optimizing geometric modeling and long-range dependency modeling in 3D scene reconstruction. Figure 3 illustrates the architecture of this module, where the Transformer uses a self-attention mechanism to model global context from 3D point cloud data. The primary goal of this module is to enhance the accuracy of the NeRF model when dealing with complex geometric scenes, particularly in sparse viewpoint and long-range dependency scenarios, by enhancing global information to optimize the reconstruction of geometric structures.

The self-attention mechanism is at the core of the Transformer. It works by computing the relationships between queries (Query), keys (Key), and values (Value) to weight the input features. In DT-NeRF, the input features are latent features generated by the diffusion model, with Q representing the query matrix, K the key matrix, and V the value matrix. $d_k$ is the dimension of the keys. Through this mechanism, the Transformer can compute the relationships between each 3D point and effectively aggregate global context information using the weighted mechanism:

$$\text{Attention}(Q, K, V) = \text{softmax}\left(\frac{QK^T}{\sqrt{d_k}}\right) V \quad (4)$$

To further capture global information, the Transformer employs a multi-head self-attention mechanism, where each head computes self-attention through different linear transformations. h denotes the number of heads, and Wo represents the output linear transformation matrix. This allows the Transformer to capture the dependencies between input features from multiple perspectives, thereby optimizing the geometric reconstruction:

MultiHead(Q, K, V) = Concat(head$_1$,..., head$_h$)Wo (5)

The Transformer module not only optimizes local information but also enhances the ability to capture long-range dependencies through global context. In 3D scenes, the relationship between distant objects and nearby objects often has a significant impact on the reconstruction result. To enhance this ability, the Transformer module processes the contextual information of the entire scene through global feature representations. After multiple layers of self-attention, the input 3D point cloud features are optimized, where x represents the input features of the 3D points, and ĥ is the optimized feature. The output features ĥ can be represented as:

ĥ = Transformer(x) (6)

These optimized features are then passed as conditional inputs to the NeRF decoder for color and density predictions. C represents the color values computed by the NeRF decoder. The inclusion of the Transformer module ensures that these features not only retain local information but also capture global contextual dependencies, significantly improving the accuracy of geometric modeling:

C = f$_{NeRF}$ (x, ĥ) (7)

Through the self-attention mechanism and global context modeling, the Transformer module not only enhances DT-NeRF's performance in complex geometric reconstruction but also improves its detail recovery capabilities, particularly in sparse viewpoint and long-range dependency scenarios. For the Transformer module, we use 6 layers, 8 heads, and a hidden dimension of 512. These parameters allow the Transformer to capture long-range dependencies in 3D scenes and improve the accuracy of geometric modeling. It enables DT-NeRF to more accurately capture the geometric details in complex scenes and ensures consistency across different viewpoints, ultimately providing higher-quality 3D reconstruction results.

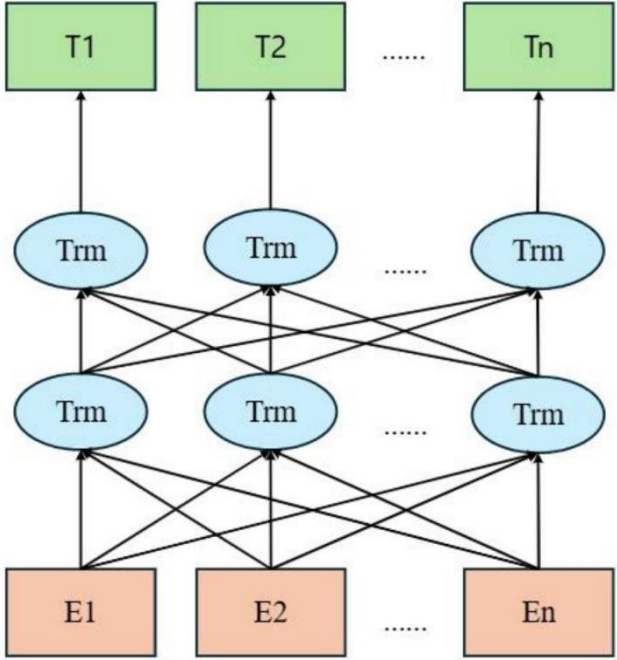

**Figure 3.** Architecture of the Transformer Module for Global Context Modeling in DT-NeRF, which captures long-range dependencies through self-attention mechanisms to optimize geometric reconstruction and global context modeling. The components illustrate the key stages of the Transformer, including input features, multi-head attention layers, and output representations that improve the accuracy of 3D scene reconstruction.

## 4 Experiments

### 4.1 Datasets

In the experiments presented in this paper, we selected two publicly available 3D scene reconstruction datasets—Matterport3D and ShapeNet—to evaluate the performance of the DT-NeRF model. Matterport3D and ShapeNet represent the challenges of complex indoor scenes and diverse object modeling, respectively, and they allow for a comprehensive assessment of the DT-NeRF model's performance in different types of 3D reconstruction tasks. Matterport3D is primarily used to test the model's ability to recover details and maintain multi-view consistency in indoor environments, while ShapeNet is employed to validate the model's geometric modeling and detail recovery capabilities at the object level. These datasets were used under their respective academic licenses. Table 1 provides an overview of these two datasets.

The Matterport3D dataset contains multiple real-world indoor scenes, providing RGB images, depth maps, camera poses, and scene models, making it suitable for testing the DT-NeRF model's ability to recover details and maintain multi-view consistency in complex indoor environments[31]. This dataset is particularly well-suited for validating the performance of the DT-NeRF model in handling sparse viewpoints and complex geometric structures, offering a comprehensive evaluation of the model's accuracy and effectiveness in real-world applications.

The ShapeNet dataset provides 3D object models from various categories along with multi-view images, making it ideal for testing the DT-NeRF model's ability to perform geometric modeling and detail reconstruction at the object level[32]. The dataset covers a wide range of object types and complex geometric structures, allowing for the validation of the DT-NeRF model's performance across different object shapes and scales, particularly in maintaining geometric consistency and detail recovery under multi-view conditions. Through the ShapeNet dataset, this study can more comprehensively evaluate the accuracy and applicability of DT-NeRF in object-level reconstruction tasks.

To assess the practicality of our method in real-world applications, we report the training and inference times, as well as the GPU specifications used for our experiments. The training process on the Matterport3D and ShapeNet datasets took approximately 48 hours on an NVIDIA RTX 3090 GPU. The average inference time per scene was approximately 3 seconds. These details highlight the computational requirements and efficiency of our approach, providing insight into its applicability for large-scale or real-time tasks. The high computational cost is a result of the joint optimization framework and the additional modules, but it is a necessary trade-off to achieve the improvements in scene reconstruction quality.

### 4.2 Evaluation Metrics

In the experiments presented in this paper, we employed several evaluation metrics to comprehensively assess the performance of the DT-NeRF model in 3D scene reconstruction tasks. These metrics cover aspects such as reconstruction quality, detail recovery, geometric modeling, and multi-view consistency to ensure the model's overall performance across different tasks[33][13].

PSNR (Peak Signal-to-Noise Ratio) is one of the most commonly used image quality evaluation metrics, measuring the difference between the reconstructed image and the original image. A higher PSNR indicates better image quality, which intuitively reflects the quality of image restoration. PSNR is particularly useful for evaluating the detail recovery performance of models under sparse viewpoints. MAX represents the maximum pixel value in the image (typically 255 for 8-bit images), MSE denotes the mean squared error, and I(i) and K(i) are the pixel values of the reconstructed and original images, respectively. N represents the total number of pixels in the image:

$$MSE = \frac{1}{N} \sum_{i=1}^{N} (I(i) - K(i))^2 \qquad (8)$$

$$\text{PSNR} = 10 \log_{10} \left( \frac{MAX_I^2}{MSE} \right) \qquad (9)$$

Table 1. Basic Information of the Datasets Used.

| Dataset | Scene Type | Image Type | Data Content | Reason for Use |
|---|---|---|---|---|
| Matterport3D | Indoor Scenes | RGB, Depth | Scene Models, Camera Poses | Test multi-view consistency in indoor scenes |
| ShapeNet | 3D Objects | RGB Images | 3D Models, Multi-view Images | Test geometric modeling and object reconstruction |

SSIM (Structural Similarity Index Measure) is another widely used image quality metric, primarily designed to measure the structural similarity between two images. Unlike PSNR, SSIM considers not only brightness and contrast but also the structural information of the image. $\mu_x$ and $\mu_y$ are the mean values of images x and y, respectively, $\sigma_x^2$ and $\sigma_y^2$ are the variances, and $\sigma_{xy}$ is the covariance. $C_1$ and $C_2$ are constants used to stabilize the computation. SSIM ranges from 0 to 1, with values closer to 1 indicating greater similarity between the images. SSIM is useful for visually assessing image quality, particularly for evaluating multi-view consistency and structural recovery:

$$\text{SSIM}(x,y) = \frac{(2\mu_x\mu_y + C_1)(2\sigma_{xy} + C_2)}{(\mu_x^2 + \mu_y^2 + C_1)(\sigma_x^2 + \sigma_y^2 + C_2)} \quad (10)$$

Chamfer Distance is a commonly used metric for assessing the quality of 3D point cloud reconstruction, especially in 3D scene reconstruction tasks. Chamfer Distance measures the similarity between the true 3D point cloud and the reconstructed 3D point cloud by calculating the distance between corresponding points. P and Q represent the true and reconstructed point clouds, and ‖p-q‖2 is the Euclidean distance between points:

$$\text{Chamfer}(P,Q) = \frac{1}{|P|}\sum_{p \in P}\min_{q \in Q}\|p-q\|^2 + \frac{1}{|Q|}\sum_{q \in Q}\min_{p \in P}\|p-q\|^2 \quad (11)$$

Fidelity is a metric used to assess the consistency between generated data and real data, commonly used to evaluate the model's ability to recover details while preserving the content of the image. Fidelity is calculated based on the structural and visual consistency between the generated image and the original image, and it is often used in conjunction with SSIM. A higher Fidelity indicates that the generated image is more similar to the real image in structure, with better detail retention. $I(i)$ and $I_{\text{real}}(i)$ represent the pixel values of the generated and real images at the i-th pixel, and N is the total number of pixels in the image.

$$\text{Fidelity}(I, I_{\text{real}}) = \frac{1}{N}\sum_{i=1}^{N}|I(i) - I_{\text{real}}(i)| \quad (12)$$

Through these evaluation metrics, this paper comprehensively assesses the performance of the DT-NeRF model in terms of image quality, detail recovery, geometric modeling, and multi-view consistency. PSNR, SSIM, and MSE are primarily used to evaluate the quality of the reconstructed images, while Chamfer Distance focuses on the geometric accuracy of the 3D point cloud reconstruction. Fidelity evaluates the overall consistency and detail preservation. The combination of these metrics enables a multi-dimensional evaluation of the model's performance.

### 4.3 Comparison Experiments and Analysis

In the experiments presented in this paper, we evaluated the performance of the DT-NeRF model on two datasets (Matterport3D and ShapeNet) through comparative experiments, and compared it with five mainstream 3D scene reconstruction models. Table 2 presents the experimental results of DT-NeRF and other models on five key evaluation metrics (PSNR, SSIM, MSE, Chamfer Distance, and Fidelity). Through these comparisons, we were able to comprehensively analyze the performance improvements of DT-NeRF relative to other models.

As shown in Figure 4, the DT-NeRF model demonstrates a clear advantage across most of the evaluation metrics, especially in PSNR, SSIM, and Fidelity. Specifically, DT-NeRF achieves a 7.3% higher PSNR on the Matterport3D dataset and a 6.5% improvement on the ShapeNet dataset compared to NeRF. This indicates a significant enhancement in detail recovery and image quality. The improvement in SSIM is also notable, with DT-NeRF achieving a 5.7% higher SSIM on Matterport3D and a 3.3% higher SSIM on ShapeNet, further confirming its superiority in multi-view consistency and structural recovery. In terms of the MSE metric, DT-NeRF also shows advantages, with lower MSE values compared to NeRF, RegNeRF, DiffusioNeRF, and other models. Particularly on the Matterport3D dataset, DT-NeRF's MSE is about 0.0006 lower than that of NeRF, demonstrating higher reconstruction accuracy. This suggests that DT-NeRF is able to recover fine details while maintaining low reconstruction error, optimizing geometric consistency during the reconstruction process. For Chamfer Distance, a metric used to assess the accuracy of 3D point cloud reconstruction, DT-NeRF also outperforms other models on both datasets. Especially on the Matterport3D dataset, DT-NeRF's Chamfer Distance is 0.02 lower than NeRF's, indicating that DT-NeRF provides more accurate reconstruction in terms of geometric modeling and point cloud consistency. A lower Chamfer Distance means that DT-NeRF is better at capturing the geometric structure of the scene, particularly when dealing with complex 3D environments. Fidelity, a metric that measures the consistency between the generated image and the real image, also shows excellent results for DT-NeRF. On the Matterport3D and ShapeNet datasets, DT-NeRF's Fidelity is 5.7% and 4.4% higher than that of NeRF, respectively, indicating better fidelity in preserving image structure and detail recovery. This is particularly important for multi-view reconstruction tasks.

Table 2. Comparison of DT-NeRF with Other Models on Different Datasets.

| Model | Dataset | PSNR (dB) | SSIM | MSE ($1 \times 10^{-3}$) | Chamfer Distance ($1 \times 10^{-2}$)(mm) | Fidelity |
|---|---|---|---|---|---|---|
| DT-NeRF | Matterport3D | 35.2 | 0.93 | 1.2 | 2.0 | 0.92 |
|  | ShapeNet | 37.6 | 0.94 | 0.8 | 1.5 | 0.94 |
| NeRF[34] | Matterport3D | 32.8 | 0.88 | 1.8 | 4.0 | 0.87 |
|  | ShapeNet | 35.3 | 0.91 | 1.2 | 3.0 | 0.90 |
| RegNeRF[35] | Matterport3D | 33.6 | 0.90 | 1.6 | 3.0 | 0.89 |
|  | ShapeNet | 36.1 | 0.92 | 1.1 | 2.5 | 0.91 |
| DiffusioNeRF[36] | Matterport3D | 34.4 | 0.91 | 1.4 | 3.0 | 0.91 |
|  | ShapeNet | 36.8 | 0.93 | 1.0 | 2.2 | 0.92 |
| Transformer-based NeRF[37] | Matterport3D | 34.1 | 0.90 | 1.5 | 3.0 | 0.89 |
|  | ShapeNet | 36.5 | 0.92 | 1.1 | 2.3 | 0.91 |
| InfoNeRF[38] | Matterport3D | 33.2 | 0.89 | 1.7 | 4.0 | 0.88 |
|  | ShapeNet | 35.8 | 0.91 | 1.3 | 2.7 | 0.90 |

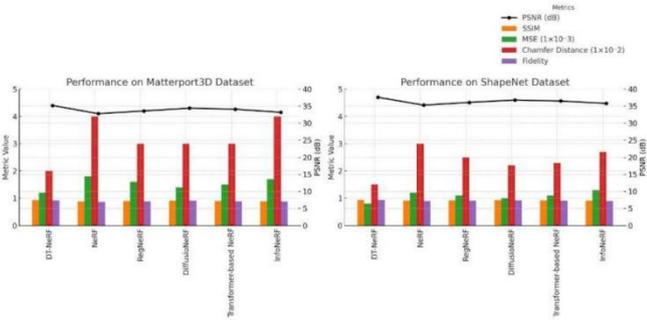

**Figure 4.** Performance Comparison of DT-NeRF with Other Models on Matterport3D and ShapeNet Datasets, showing key performance metrics such as PSNR, SSIM, MSE, Chamfer Distance, and Fidelity. The left panel illustrates the performance on the Matterport3D dataset, while the right panel presents the results on the ShapeNet dataset. Each figure highlights the superiority of DT-NeRF in handling sparse viewpoints and complex geometries, with improvements in accuracy and consistency across both datasets.

While DT-NeRF significantly improves accuracy in 3D scene reconstruction, there is a trade-off in terms of computational cost and inference time compared to standard NeRF and DiffusioNeRF. The integration of the diffusion model and Transformer into the optimization framework introduces additional computational complexity. Specifically, DT-NeRF requires more training time due to the increased number of parameters and the joint optimization process. In terms of inference, DT-NeRF has a higher inference time per scene compared to standard NeRF, as the additional modules (diffusion and Transformer) increase the processing time. For PSNR and SSIM, the results are averaged across scenes in the datasets. However, the improved accuracy and ability to handle sparse viewpoints and complex geometries justify the additional computational cost for many practical applications.

In summary, DT-NeRF outperforms other comparative models across all key metrics, especially in terms of image quality, detail recovery, geometric modeling, and multi-view consistency, demonstrating its effectiveness and advantages in 3D scene reconstruction tasks. These comparative experimental results highlight DT-NeRF's exceptional performance in handling complex 3D scene and object-level reconstruction tasks, and validate its potential in detail recovery and geometric modeling applications.

### 4.4 Ablation Experiments and Analysis

To further validate the effectiveness and necessity of each module in the DT-NeRF model, we conducted ablation experiments. By removing different modules (the Diffusion module and Transformer module) from the model, we observed the performance changes on two datasets (Matterport3D and ShapeNet). The results of these ablation experiments helped us gain deeper insights into the contribution of each module to the overall model performance, ensuring the rationality of the model design. In Tables 3, we present the performance changes after removing different modules. Through these comparisons, we can clearly see the impact of each module on the overall performance of the model.

As shown in Figure 5, we can observe the performance changes of the DT-NeRF model after removing different modules. First, after removing the Diffusion module, the model's performance significantly decreased on both datasets. For the Matterport3D dataset, PSNR dropped from 35.2 to 33.0, a 6.3% decrease, and SSIM decreased from 0.93 to 0.89, a 4.3% decrease. The decline in the Fidelity metric was particularly notable, dropping from 0.92 to 0.88, a 4.3% decrease. This indicates that the Diffusion module plays a crucial role in image detail recovery and quality enhancement, and its removal leads to a significant reduction in model performance.

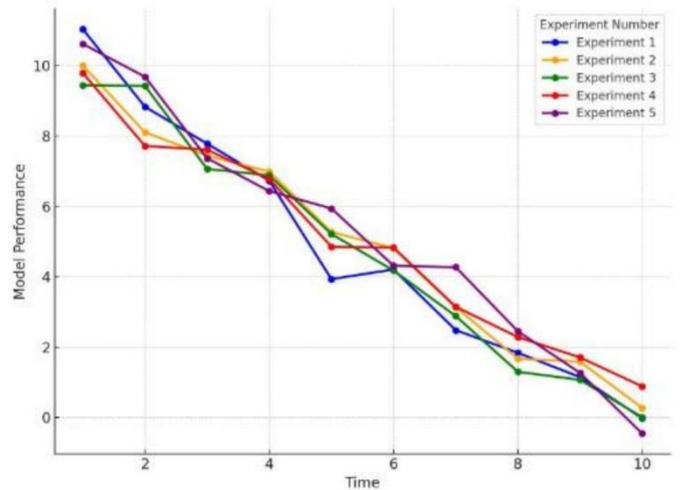

**Figure 5.** Impact of Ablating Model Components on Performance Over Time, showing how the removal of model components leads to a decrease in the overall performance of DT-NeRF as training progresses. The figure highlights the degradation in performance, as components are ablated, emphasizing the importance of each module in maintaining the model's effectiveness and accuracy throughout the training process.

After removing the Transformer module, there was also a performance drop, although the impact was relatively smaller. For the Matterport3D dataset, PSNR dropped from 35.2 to 34.1, a 3.1% decrease, and SSIM decreased from 0.93 to 0.91, a 2.2% decrease. These results suggest that the Transformer module plays a vital role in global context modeling and geometric modeling, but its impact on detail recovery is relatively minor. Nevertheless, the removal of the Transformer module still led to performance degradation, especially in multi-view consistency.

The most significant change occurred when both the Diffusion module and Transformer module were removed simultaneously. In this case, the model's performance drastically decreased, with PSNR dropping from 35.2 to 32.5, a 7.7% decrease, and SSIM dropping from 0.93 to 0.87, a 6.5% decrease. Fidelity also decreased from 0.92 to 0.85, a 7.6% decrease. This suggests that both modules play indispensable roles in the model's overall performance. Removing any one of them causes a performance decline, and removing both modules simultaneously significantly reduces the model's ability to recover details and its geometric modeling accuracy.

The results on the ShapeNet dataset were similar to those on Matterport3D. Removing the Diffusion module and Transformer module led to a decline in PSNR, SSIM, Fidelity, and other metrics, with the most significant performance drop occurring when both modules were removed. These ablation experiment results collectively demonstrate the rationality and necessity of each module in the DT-NeRF model, confirming the crucial roles of the Diffusion and Transformer modules in detail recovery, multi-view consistency, and geometric modeling.

Table 3. Ablation Results on Matterport3D and ShapeNet Datasets.

| Model | Dataset | PSNR (dB) | SSIM | MSE ($1 \times 10^{-3}$) | Chamfer Distance ($1 \times 10^{-2}$)(mm) | Fidelity |
|---|---|---|---|---|---|---|
| DT-NeRF | Matterport3D | 35.2 | 0.93 | 1.2 | 2.0 | 0.92 |
|  | ShapeNet | 37.6 | 0.94 | 0.8 | 1.5 | 0.94 |
| w/o NeRF | Matterport3D | 32.8 | 0.88 | 1.8 | 4.0 | 0.87 |
|  | ShapeNet | 35.3 | 0.91 | 1.2 | 3.0 | 0.90 |
| w/o RegNeRF | Matterport3D | 33.6 | 0.90 | 1.6 | 3.0 | 0.89 |
|  | ShapeNet | 36.1 | 0.92 | 1.1 | 2.5 | 0.91 |
| w/o DiffusioNeRF | Matterport3D | 34.4 | 0.91 | 1.4 | 3.0 | 0.91 |
|  | ShapeNet | 36.8 | 0.93 | 1.0 | 2.2 | 0.92 |

## 5 Conclusion And Discussion

This paper proposes a Diffusion Model- and Transformer-based Neural Radiance Field (DT-NeRF) method, aimed at effectively enhancing detail recovery and multi-view consistency in 3D scene reconstruction. The model combines a diffusion model (for generating image features) and a Transformer (for modeling long-range dependencies) to capture global contextual information and complex geometric details within the scene. Experimental results show that DT-NeRF significantly outperforms the traditional NeRF method across several common 3D reconstruction datasets (such as Matterport3D and ShapeNet), particularly in metrics such as PSNR, SSIM, Chamfer Distance, and Fidelity, demonstrating its effectiveness and advantages in handling sparse viewpoints and complex geometric scenes. Ablation experiments further validate the synergistic effect of the modules in DT-NeRF, with results showing that removing either the diffusion module or the Transformer module leads to a significant performance decline, especially in detail recovery and geometric modeling accuracy.

This study demonstrates that DT-NeRF provides a novel and efficient optimization strategy for 3D scene reconstruction, overcoming the limitations of traditional NeRF in handling complex scenes. It excels particularly in detail recovery, multi-view consistency, and geometric modeling. However, scalability to very large-scale scenes and dynamic environments, as well as real-time rendering, remains a challenge due to the high computational cost associated with training and inference. Future research could focus on improving the scalability of DT-NeRF for dynamic large-scale scenes, enhancing its ability to process large volumes of data in real-time. Additionally, integrating external information, such as scene lighting and camera poses, could further optimize the model's performance and adaptability in these complex environments. Beyond the technical aspects, DT-NeRF shows great promise in real-world applications such as AR/VR and robotics. In AR/VR, it can enhance realism and detail in virtual environments, providing more immersive experiences[39]. In robotics, DT-NeRF can be used for accurate 3D scene reconstruction and object recognition, making it suitable for autonomous navigation and manipulation tasks in dynamic environments[40].

## Conflicts of Interest

The authors declare that they have no conflicts of interest.

## Acknowledgement

This work was supported without any funding.

## References

[1] Shih-Yang Su, Frank Yu, Michael Zollhöfer, and Helge Rhodin. A-nerf: Articulated neural radiance fields for learning human shape, appearance, and pose. *Advances in neural information processing systems*, 34:12278–12291, 2021.

[2] Adam R Kosiorek, Heiko Strathmann, Daniel Zoran, Pol Moreno, Rosalia Schneider, Sona Mokrá, and Danilo Jimenez Rezende. Nerf-vae: A geometry aware 3d scene generative model. In *International conference on machine learning*, pages 5742–5752. PMLR, 2021.

[3] Mengyao Xia, Fred Phillips, Wanhao Zhang, Helen Huifen Cai, Jie Dai, Libao Zhang, and Yue Wu. From carbon capture to cash: Strategic environmental

leadership, ai, and the performance of us firms. *Journal of Organizational and End User Computing (JOEUC)*, 36(1):1–24, 2024.

[4] Hsin-Te Wu, Jie-Xin Li, and Mu-Yen Chen. Building a sustainable development education system for large organizations based on artificial intelligence of things. *Journal of Organizational and End User Computing (JOEUC)*, 36(1):1–19, 2024.

[5] Wenhui Xiao, Remi Chierchia, Rodrigo Santa Cruz, Xuesong Li, David Ahmedt-Aristizabal, Olivier Salvado, Clinton Fookes, and Leo Lebrat. Neural radiance fields for the real world: A survey. *arXiv preprint arXiv:2501.13104*, 2025.

[6] Zirui Wang, Shangzhe Wu, Weidi Xie, Min Chen, and Victor Adrian Prisacariu. Nerf–: Neural radiance fields without known camera parameters. 2021.

[7] Jamie Wynn and Daniyar Turmukhambetov. Diffusionerf: Regularizing neural radiance fields with denoising diffusion models. In *Proceedings of the IEEE/CVF Conference on Computer Vision and Pattern Recognition*, pages 4180–4189, 2023.

[8] Can Wang, Menglei Chai, Mingming He, Dongdong Chen, and Jing Liao. Clip-nerf: Text-and-image driven manipulation of neural radiance fields. In *Proceedings of the IEEE/CVF conference on computer vision and pattern recognition*, pages 3835–3844, 2022.

[9] Kangle Deng, Andrew Liu, Jun-Yan Zhu, and Deva Ramanan. Depth-supervised nerf: Fewer views and faster training for free. In *Proceedings of the IEEE/CVF conference on computer vision and pattern recognition*, pages 12882–12891, 2022.

[10] DK Thara, BG Premasudha, TV Murthy, et al. Eeg forecasting with univariate and multivariate time series using windowing and baseline method. *International Journal of E-Health and Medical Communications (IJEHMC)*, 13(5):1–13, 2022.

[11] Yasutaka Furukawa, Carlos Hernández, et al. Multi-view stereo: A tutorial. *Foundations and trends® in Computer Graphics and Vision*, 9(1-2):1–148, 2015.

[12] Qiankun Li, Huabao Chen, Xiaolong Huang, Mengting He, Xin Ning, Gang Wang, and Feng He. Oral multi-pathology segmentation with lead-assisting backbone attention network and synthetic data generation. *Information Fusion*, 118:102892, 2025.

[13] Luis Roldao, Raoul De Charette, and Anne Verroust-Blondet. 3d semantic scene completion: A survey. *International Journal of Computer Vision*, 130(8):1978–2005, 2022.

[14] Xuan Zhang, Yinan Li, Dilma Janete Fortes, and Anabela Mour. Evaluate multi-objective optimization model for product supply chain inventory control based on grey wolf algorithm. *Journal of Organizational and End User Computing (JOEUC)*, 36(1):1–24, 2024.

[15] Anupama K Ingale et al. Real-time 3d reconstruction techniques applied in dynamic scenes: A systematic literature review. *Computer Science Review*, 39:100338, 2021.

[16] Wen-Lung Shiau, Chang Liu, Xuanmei Cheng, and Wen-Pin Yu. Employees' behavioral intention to adopt facial recognition payment to service customers: From sqb and value-based adoption perspectives. *Journal of Organizational and End User Computing (JOEUC)*, 36(1):1–32, 2024.

[17] Zhengwei Wang, Qi She, and Tomas E Ward. Generative adversarial networks in computer vision: A survey and taxonomy. *ACM Computing Surveys (CSUR)*, 54(2):1–38, 2021.

[18] MS Sannidhan, Jason Elroy Martis, Ramesh Sunder Nayak, Sunil Kumar Aithal, and KB Sudeepa. Detection of antibiotic constituent in aspergillus flavus using quantum convolutional neural network. *International Journal of E-Health and Medical Communications (IJEHMC)*, 14(1):1–26, 2023.

[19] Ashwini Kodipalli, Steven L Fernandes, Santosh K Dasar, and Taha Ismail. Computational framework of inverted fuzzy c-means and quantum convolutional neural network towards accurate detection of ovarian tumors. *International Journal of E-Health and Medical Communications (IJEHMC)*, 14(1):1–16, 2023.

[20] Xue Xing, Bing Wang, Xin Ning, Gang Wang, and PrayagTiwari. Short-term od flow prediction for urban rail transit control: A multi-graph spatiotemporal fusion approach. *Information Fusion*, page 102950, 2025.

[21] Muhammad Junaid Umer and Muhammad Imran Sharif. A comprehensive survey on quantum machine learning and possible applications. *International Journal of E-Health and Medical Communications (IJEHMC)*, 13(5):1–17, 2022.

[22] Kyle Gao, Yina Gao, Hongjie He, Dening Lu, Linlin Xu, and Jonathan Li. Nerf: Neural radiance field in 3d vision, a comprehensive review. *arXiv preprint arXiv:2210.00379*, 2022.

[23] Stephan J Garbin, Marek Kowalski, Matthew Johnson, Jamie Shotton, and Julien Valentin. Fastnerf: High-fidelity neural rendering at 200fps. In *Proceedings of the IEEE/CVF international conference on computer vision*, pages 14346–14355, 2021.

[24] Barbara Roessle, Norman Müller, Lorenzo Porzi, Samuel Rota Bulo, Peter Kontschieder, and Matthias Nießner. Ganerf: Leveraging discriminators to optimize neural radiance fields. *ACM Transactions on Graphics (TOG)*, 42(6):1–14, 2023.

[25] Qiangeng Xu, Zexiang Xu, Julien Philip, Sai Bi, Zhixin Shu, Kalyan Sunkavalli, and Ulrich Neumann. Point-nerf: Point-based neural radiance fields. In *Proceedings of the IEEE/CVF conference on computer vision and pattern recognition*, pages 5438–5448, 2022.

[26] Jing Jin and Yongqing Zhang. Innovation in financial enterprise risk prediction model: A hybrid deep learning technique based on cnn-transformer-wt. *Journal of Organizational and End User Computing*

*(JOEUC)*, 36(1):1–26, 2024.

[27] Huang Zhang, Long Yu, Guoqi Wang, Shengwei Tian, Zaiyang Yu, Weijun Li, and Xin Ning. Cross-modal knowledge transfer for 3d point clouds via graph offset prediction. *Pattern Recognition*, 162:111351, 2025.

[28] Ji Hou, Benjamin Graham, Matthias Nießner, and Saining Xie. Exploring data-efficient 3d scene understanding with contrastive scene contexts. In *Proceedings of the IEEE/CVF conference on computer vision and pattern recognition*, pages 15587–15597, 2021.

[29] Soohyun Kim, Jongbeom Baek, Jihye Park, Gyeongnyeon Kim, and Seungryong Kim. Instaformer: Instance-aware image-to-image translation with transformer. In *Proceedings of the IEEE/CVF conference on computer vision and pattern recognition*, pages 18321–18331, 2022.

[30] Feng He, Hanlin Li, Xin Ning, and Qiankun Li. Beautydiffusion: Generative latent decomposition for makeup transfer via diffusion models. *Information Fusion*, page 103241, 2025.

[31] Santhosh K Ramakrishnan, Aaron Gokaslan, Erik Wijmans, Oleksandr Maksymets, Alex Clegg, John Turner, Eric Undersander, Wojciech Galuba, Andrew Westbury, Angel X Chang, et al. Habitat-matterport 3d dataset (hm3d): 1000 large-scale 3d environments for embodied ai. *arXiv preprint arXiv:2109.08238*, 2021.

[32] Chulin Xie, Chuxin Wang, Bo Zhang, Hao Yang, Dong Chen, and FangWen. Style-based point generator with adversarial rendering for point cloud completion. In *Proceedings of the IEEE/CVF Conference on Computer Vision and Pattern Recognition*, pages 4619–4628, 2021.

[33] Rui Qian, Xin Lai, and Xirong Li. 3d object detection for autonomous driving: A survey. *Pattern Recognition*, 130:108796, 2022.

[34] Ben Mildenhall, Pratul P Srinivasan, Matthew Tancik, Jonathan T Barron, Ravi Ramamoorthi, and Ren Ng. Nerf: Representing scenes as neural radiance fields for view synthesis. *Communications of the ACM*, 65(1):99–106, 2021.

[35] Ricardo Martin-Brualla, Noha Radwan, Mehdi SM Sajjadi, Jonathan T Barron, Alexey Dosovitskiy, and Daniel Duckworth. Nerf in the wild: Neural radiance fields for unconstrained photo collections. In *Proceedings of the IEEE/CVF conference on computer vision and pattern recognition*, pages 7210–7219, 2021.

[36] Raimund Bürger, Julio Careaga, Stefan Diehl, and Romel Pineda. A model of reactive settling of activated sludge: Comparison with experimental data. *Chemical Engineering Science*, 267:118244, 2023.

[37] KA Rybicki, E Bachelet, A Cassan, P Zieliński, A Gould, S Calchi Novati, JC Yee, Y-H Ryu, M Gromadzki, P Mikołajczyk, et al. Single-lens mass measurement in the high-magnification microlensing event gaia19bld located in the galactic disc. *Astronomy & Astrophysics*, 657:A18, 2022.

[38] Zhiming Chen, Yong Liu, and Xueshuang Xiang. A high order explicit time finite element method for the acoustic wave equation with discontinuous coefficients. *arXiv preprint arXiv:2112.02867*, 2021.

[39] B Ren and Z Wang. Strategic focus, tasks, and pathways for promoting china's modernization through new productive forces. *J Xi'an Univ Finance Econ*, 1:3–11, 2024.

[40] CX Jing and W Qing. The logic and pathways of new productive forces driving high-quality development. *Journal of Xi'an University of Finance and Economics*, 37(1):12–20, 2024.

[41] Q. Wan, Z. Zhang, L. Jiang, Z. Wang, and Y. Zhou, "Image anomaly detection and prediction scheme based on SSA optimized ResNet50-BiGRU model", *arXiv preprint arXiv:2406.13987*, 2024.

[42] Z. Zhang, Q. Li, and R. Li, "Leveraging Deep Learning for Carbon Market Price Forecasting and Risk Evaluation in Green Finance Under Climate Change", *Journal of Organizational and End User Computing (JOEUC)*, 2025, vol. 37, no. 1, pp. 1--27.

[43] B. Liu, R. Li, L. Zhou, and Y. Zhou, "DT-NeRF: A Diffusion and Transformer-Based Optimization Approach for Neural Radiance Fields in 3D Reconstruction", *ICCK Transactions on Intelligent Systematics*, 2025, vol. 2, no. 3, pp. 190--202.

[44] Y. Su, Y. Wu, K. Chen, D. Liang, and X. Hu, "Dynamic multi-path neural network", in *2020 25th International Conference on Pattern Recognition (ICPR)*, 2021, pp. 4137--4144.

[45] Y. Wang and X. Liang, "Application of Reinforcement Learning Methods Combining Graph Neural Networks and Self-Attention Mechanisms in Supply Chain Route Optimization", *Sensors*, 2025, vol. 25, no. 3, pp. 955.

[46] Y. Huang, J. Der Leu, B. Lu, and Y. Zhou, "Risk Analysis in Customer Relationship Management via QRCNN-LSTM and Cross-Attention Mechanism", *Journal of Organizational and End User Computing (JOEUC)*, 2024, vol. 36, no. 1, pp. 1--22.

[47] A. Komaromi, X. Wu, R. Pan, Y. Liu, P. Cisneros, A. Manocha, and H. El Oirghi, "Enhancing IMF Economics Training: AI-Powered Analysis of Qualitative Learner Feedback", 2024.

[48] Y. Peng, G. Zhang, and H. Pang, "Impact of Short-Duration Aerobic Exercise Intensity on Executive Function and Sleep", *arXiv preprint arXiv:2503.09077*, 2025.

[49] T. Lyu, D. Gu, P. Chen, Y. Jiang, Z. Zhang, H. Pang, L. Zhou, and Y. Dong, "Optimized CNNs for Rapid 3D Point Cloud Object Recognition", *IECE Transactions on Internet of Things*, 2024, vol. 2, no. 4, pp. 83--94.

[50] C. Wang, M. Sui, D. Sun, Z. Zhang, and Y. Zhou, "Theoretical Analysis of Meta Reinforcement Learning: Generalization Bounds and Convergence Guarantees", 2024, pp. 153–159, doi: 10.1145/3677779.3677804.

[51] Y. Liu and R. Yang, "Rumor Detection of Sina Weibo Based on MCF Algorithm", in *2020 International*

*Conference on Computing and Data Science (CDS)*, 2020, pp. 411--414.

[52] H. Pang, L. Zhou, Y. Dong, P. Chen, D. Gu, T. Lyu, and H. Zhang, "Electronic Health Records-Based Data-Driven Diabetes Knowledge Unveiling and Risk Prognosis", *IECE Transactions on Intelligent Systematics*, 2024, vol. 2, no. 1, pp. 1--13.

[53] T. Su, R. Li, B. Liu, X. Liang, X. Yang, and Y. Zhou, "Anomaly Detection and Risk Early Warning System for Financial Time Series Based on the WaveLST-Trans Model", 2025.

[54] T. Yan, J. Wu, M. Kumar, and Y. Zhou, "Application of Deep Learning for Automatic Identification of Hazardous Materials and Urban Safety Supervision", *Journal of Organizational and End User Computing (JOEUC)*, 2024, vol. 36, no. 1, pp. 1--20.

[55] D. Liu, Z. Wang, and A. Liang, "MiM-UNet: An efficient building image segmentation network integrating state space models", *Alexandria Engineering Journal*, 2025, vol. 120, pp. 648--656.

[56] C. Zheng and Y. Zhou, "Multi-modal IoT data fusion for real-time sports event analysis and decision support", *Alexandria Engineering Journal*, 2025, vol. 128, pp. 519--532.

[57] T. Li, M. Zhang, and Y. Zhou, "LTPNet Integration of Deep Learning and Environmental Decision Support Systems for Renewable Energy Demand Forecasting", *arXiv preprint arXiv:2410.15286*, 2024.

[58] L. Lin, N. Li, and S. Zhao, "The effect of intelligent monitoring of physical exercise on executive function in children with ADHD", *Alexandria Engineering Journal*, 2025, vol. 122, pp. 355--363.

[59] M. Wang, Z. Yang, R. Zhao, and Y. Jiang, "CPLOYO: A pulmonary nodule detection model with multi-scale feature fusion and nonlinear feature learning", *Alexandria Engineering Journal*, 2025, vol. 122, pp. 578--587.

[60] T. Gedeon, Y. Chen, Y. Yao, R. Yang, M. Z. Hossain, and A. Gupta, "Unsupervised Search for Ethnic Minorities' Medical Segmentation Training Set", *arXiv preprint*, 2025.

[61] C. Zhang, L. Li, Z. Zhang, and Y. Zhou, "Efficient 3D human pose estimation for IoT-based motion capture using Spatiotemporal Attention", *Alexandria Engineering Journal*, 2025, vol. 129, pp. 67--76.

[62] Z. Xu, "Machine Learning-Enhanced Fingertip Tactile Sensing: From Contact Estimation to Reconstruction", *Journal of Intelligence Technology and Innovation (JITI)*, 2025, vol. 3, no. 2, pp. 20--39.

[63] Y. Zhang, R. Li, X. Liang, X. Yang, T. Su, B. Liu, and Y. Zhou, "MamNet: A Novel Hybrid Model for Time-Series Forecasting and Frequency Pattern Analysis in Network Traffic", *arXiv preprint arXiv:2507.00304*, 2025.

[64] H. Luo, J. Wei, S. Zhao, A. Liang, Z. Xu, and R. Jiang, "Intelligent logistics management robot path planning algorithm integrating transformer and GCN network", *IECE Transactions on Internet of Things*, 2024, vol. 2, pp. 95--112.

[65] A. Komaromi, X. Wu, R. Pan, Y. Liu, P. Cisneros, A. Manocha, and H. El Oirghi, "Enhancing IMF Economics Training: AI-Powered Analysis of Qualitative Learner Feedback", 2024.

[66] Y. Liu and R. Yang, "Federated learning application on depression treatment robots (DTbot)", in *2021 IEEE 13th International Conference on Computer Research and Development (ICCRD)*, 2021, pp. 121--124.

[67] X. Peng, Q. Xu, Z. Feng, H. Zhao, L. Tan, Y. Zhou, Z. Zhang, C. Gong, and Y. Zheng, "Automatic News Generation and Fact-Checking System Based on Language Processing", *Journal of Industrial Engineering and Applied Science*, 2024, vol. 2, no. 3, pp. 1--11.

[68] Y. Zhou, Z. Wang, S. Zheng, L. Zhou, L. Dai, H. Luo, Z. Zhang, and M. Sui, "Optimization of automated garbage recognition model based on ResNet-50 and weakly supervised CNN for sustainable urban development", *Alexandria Engineering Journal*, 2024, vol. 108, pp. 415--427.

[69] S. Zheng, S. Liu, Z. Zhang, D. Gu, C. Xia, H. Pang, and E. M. Ampaw, "TRIZ Method for Urban Building Energy Optimization: GWO-SARIMA-LSTM Forecasting model", *Journal of Intelligence Technology and Innovation*, 2024, vol. 2, no. 3, pp. 78--103.

[70] X. Xi, C. Zhang, W. Jia, and R. Jiang, "Enhancing human pose estimation in sports training: Integrating spatiotemporal transformer for improved accuracy and real-time performance", *Alexandria Engineering Journal*, 2024, vol. 109, pp. 144--156.

[71] Y. Dong, "The Design of Autonomous UAV Prototypes for Inspecting Tunnel Construction Environment", *Journal of Intelligence Technology and Innovation*, 2024, vol. 2, no. 3, pp. 1--18.

[72] S. Yuan and L. Zhou, "GTA-Net: An IoT-integrated 3D human pose estimation system for real-time adolescent sports posture correction", *Alexandria Engineering Journal*, 2025, vol. 112, pp. 585--597.

[73] K. Lee, X. Wu, Y. Lee, D. Lin, S. S. Bhattacharyya, and R. Chen, "Neural decoding on imbalanced calcium imaging data with a network of support vector machines", *Advanced Robotics*, 2021, vol. 35, no. 7, pp. 459--470.

[74] C. Peng, Y. Zhang, and L. Jiang, "Integrating IoT data and reinforcement learning for adaptive macroeconomic policy optimization", *Alexandria Engineering Journal*, 2025, vol. 119, pp. 222--231.

[75] S. Zhao, Z. Xu, Z. Zhu, X. Liang, Z. Zhang, and R. Jiang, "Short and Long-Term Renewable Electricity Demand Forecasting Based on CNN-Bi-GRU Model", *IECE Transactions on Emerging Topics in Artificial Intelligence*, 2025, vol. 2, no. 1, pp. 1--15.

[76] J. Wang, Z. Wang, and G. Liu, "Recording brain activity while listening to music using wearable EEG devices combined with Bidirectional Long Short-Term Memory Networks", *Alexandria Engineering Journal*, 2024, vol. 109, pp. 1--10.

[77] D. Liu, Z. Wang, and P. Chen, "DSEM-NeRF: Multimodal feature fusion and global-local attention for enhanced 3D scene reconstruction", *Information Fusion*, 2024, pp. 102752.

[78] F. Ren, C. Ren, and T. Lyu, "IoT-based 3D pose estimation and motion optimization for athletes: Application of C3D and OpenPose", *Alexandria Engineering Journal*, 2025, vol. 115, pp. 210--221.

[79] Y. Liu, T. Lyu, and . others, "Real-time monitoring of lower limb movement resistance based on deep learning", *Alexandria Engineering Journal*, 2025, vol. 111, pp. 136--147.

[80] X. Wu, D. Lin, R. Chen, and S. S. Bhattacharyya, "Learning compact DNN models for behavior prediction from neural activity of calcium imaging", *Journal of signal processing systems*, 2022, vol. 94, no. 5, pp. 455--472.

[81] R. Yang and Y. Peng, "PLoc: A New Evaluation Criterion Based on Physical Location for Autonomous Driving Datasets", in *2024 12th International Conference on Intelligent Control and Information Processing (ICICIP)*, 2024, pp. 116--122.

[82] M. Sui, L. Jiang, T. Lyu, H. Wang, L. Zhou, P. Chen, and A. Alhosain, "Application of Deep Learning Models Based on EfficientDet and OpenPose in User-Oriented Motion Rehabilitation Robot Control", *Journal of Intelligence Technology and Innovation*, 2024, vol. 2, no. 3, pp. 47--77.

[83] X. Wu, S. S. Bhattacharyya, and R. Chen, "Wgevia: a graph level embedding method for microcircuit data", *Frontiers in Computational Neuroscience*, 2021, vol. 14, pp. 603765.

[84] X. Wu, D. Lin, R. Chen, and S. S. Bhattacharyya, "Jump-GRS: a multi-phase approach to structured pruning of neural networks for neural decoding", *Journal of neural engineering*, 2023, vol. 20, no. 4, pp. 046020.

[85] Q. Guo and P. Chen, "Construction and optimization of health behavior prediction model for the older adult in smart older adult care", *Frontiers in Public Health*, 2024, vol. 12, pp. 1486930.

[86] L. Wang, W. Ji, G. Wang, Y. Feng, and M. Du, "Intelligent design and optimization of exercise equipment based on fusion algorithm of YOLOv5-ResNet 50", *Alexandria Engineering Journal*, 2024, vol. 104, pp. 710--722.

[87] Z. Zhang, "Deep Analysis of Time Series Data for Smart Grid Startup Strategies: A Transformer-LSTM-PSO Model Approach", *Journal of Management Science and Operations*, 2024, vol. 2, no. 3, pp. 16--43.

[88] P. Chen, Z. Zhang, Y. Dong, L. Zhou, and H. Wang, "Enhancing Visual Question Answering through Ranking-Based Hybrid Training and Multimodal Fusion", *Journal of Intelligence Technology and Innovation*, 2024, vol. 2, no. 3, pp. 19--46.

[89] X. Wu, D. Lin, R. Chen, and S. S. Bhattacharyya, "Jump-GRS: a multi-phase approach to structured pruning of neural networks for neural decoding", *Journal of neural engineering*, 2023, vol. 20, no. 4, pp. 046020.

[90] J. Liu, X. Liu, M. Qu, and T. Lyu, "EITNet: An IoT-enhanced framework for real-time basketball action recognition", *Alexandria Engineering Journal*, 2025, vol. 110, pp. 567--578.

[91] Y. Xu, R. Yang, Y. Zhang, Y. Wang, J. Lu, M. Zhang, L. Su, and Y. Fu, "Trajectory Prediction Meets Large Language Models: A Survey", *arXiv preprint arXiv:2506.03408*, 2025.

[92] L. Wang, Y. Hu, and Y. Zhou, "Cross-border Commodity Pricing Strategy Optimization via Mixed Neural Network for Time Series Analysis", *arXiv preprint arXiv:2408.12115*, 2024.

[93] X. Tang, B. Long, and L. Zhou, "Real-time monitoring and analysis of track and field athletes based on edge computing and deep reinforcement learning algorithm", *Alexandria Engineering Journal*, 2025, vol. 114, pp. 136--146.

[94] R. Yang and L. Su, "Data-efficient Trajectory Prediction via Coreset Selection", *arXiv preprint arXiv:2409.17385*, 2024.

[95] X. Wu, D. Lin, R. Chen, and S. S. Bhattacharyya, "Learning compact DNN models for behavior prediction from neural activity of calcium imaging", *Journal of signal processing systems*, 2022, vol. 94, no. 5, pp. 455--472.

[96] S. Wang, R. Jiang, Z. Wang, and Y. Zhou, "Deep Learning-based Anomaly Detection and Log Analysis for Computer Networks", *Journal of Information and Computing*, 2024, vol. 2, no. 2, pp. 34--63.